\documentclass[a4paper]{article}
\usepackage{times}
\usepackage[dvips]{graphicx}
\usepackage{epsfig}
\usepackage{amssymb}

\usepackage{amsmath}
\usepackage{latexsym}
\usepackage{verbatim}

\usepackage{subfigure}
\usepackage{algorithm}
\usepackage{algorithmic}

\newcommand{\OPRAm}{${\cal OPRA}_m$}
\newcommand{\EOPRAm}{${\cal EOPRA}_m$}

\begin{document}

\newenvironment{Proof}{\textit{Proof.}}{\hspace*{\fill}$\Box$}
\newenvironment{ProofSketch}{\textit{Proof Sketch.}}{\hspace*{\fill}$\Box$}

\title{
Extending Binary 
Qualitative Direction Calculi
with a Granular Distance Concept:\\
Hidden Feature Attachment
}

\author{Reinhard Moratz\\
Spatial Information Science
and Engineering\\
National Center for Geographic Information and Analysis (NCGIA)\\
University of Maine\\
5711 Boardman Hall\\
Orono, ME 04469, USA}

\bibliographystyle{plain}

\maketitle

\begin{abstract}
In this paper we introduce 
a method for 
extending binary 
qualitative direction calculi with adjustable granularity like \OPRAm or the star calculus
with a granular distance concept.
This method is similar to the concept of extending points with an internal
reference direction to get oriented points which are the basic entities
in the \OPRAm calculus. Even if the spatial objects are from a geometrical
point of view infinitesimal small points locally available reference
measures are attached. In the case of \OPRAm, a reference direction is attached.
The same principle works also with local reference distances which are called elevations.
The principle of attaching references features to a point is called hidden feature
attachment.
\end{abstract}

\section{Introduction}
A {\it qualitative} representation provides mechanisms
which characterize central essential properties of objects or configurations. 
A {\it quantitative} representation establishes 
a measure in relation to
a unit of measurement which has to be generally available.
Qualitative spatial calculi usually deal with elementary objects (e.g.,
positions, directions, regions) and qualitative relations between them
(e.g., ''adjacent'', ''on the left of'', ''included in'').

The constant general availability of common measures is now self evident. 
However, one needs only remember the example of the history of technologies of 
measurement of length to see that the more local relative measures, 
which are qualitatively represented, 
(for example, ''one piece of material is longer than another'' versus
''this thing is two meters long'') 
can be managed by biological/epigenetic cognitive systems
much more easily as absolute quantitative representations.
Typically, in Qualitative Spatial Reasoning relatively coarse distinctions
between configurations are made only.
The two main trends in Qualitative Spatial
Reasoning are topological reasoning about regions
\cite{Randell92_RCCb,renz99_RCC_Complexity,egenhofer9intersection}
and positional
reasoning about point configurations
\cite{freksa92b,frank91,Ligozat98_CardDir,DBLP:conf/ecai/Moratz06,Renz04_QDCArbGranu}.

Applications exist in which finer qualitative acceptance areas are helpful. 
The possibility to use finer qualitative distinctions can be viewed
as a stepwise transition to quantitative knowledge. The idea of using context 
dependant direction and distance intervals for the representation of spatial 
knowledge can be 
traced back to Clementini, di Felice, and Hernandez  \cite{hernandez_aij}.  
However, only special cases of reasoning were considered in their work.
Here, we will propose calculi that make direct use of 
general purpose constraint propagation.

In this paper we introduce 
a method for 
extending binary 
qualitative direction calculi with adjustable granularity like \OPRAm \cite{DBLP:conf/ecai/Moratz06}
or the star calculus \cite{Renz04_QDCArbGranu}
with a granular distance concept.
This method is similar to the concept of extending points with an internal
reference direction to get oriented points which are the basic entities
in the \OPRAm calculus \cite{DBLP:conf/ecai/Moratz06}. Even if the spatial objects are from a geometrical
point of view infinitesimal small points locally available reference
measures are attached. In the case of \OPRAm, a reference direction is attached.
The same principle works also with local reference distances which are called elevations.
The principle of attaching references features to a point is called hidden feature
attachment.

The newly proposed calculi are an extension of the \OPRAm calculus called \EOPRAm and
an extension of the star calculus. The key principle to derive these
new calculi is presented in the next section. 

\section{Hidden Feature Attachment with Point Calculi and Elevation of Points}

There is a motivation about the use of oriented points in qualitative calculi. 
This motivation is neccessary since qualitative calculi
characterize central essential properties of objects or configurations (see section \ref{introduction}).
A simple featureless point naturally would not have an internal reference direction.
The conception which makes the internal reference direction plausible is 
a transition from 
a
calculus which is based on straight line segments (dipoles) \cite{Moratz00_QSRwithLineSegs}
with
concrete length to line segments with infinitely small length. In this conceptualization the length
of the objects no longer has any importance.
Thus, only the direction of the objects is modeled.
{\em O-points}, our term for oriented points, are specified as pair of a point and a direction
on the 2D-plane \cite{DBLP:conf/ecai/Moratz06}. 

This method attaches a feature which is used as a local reference to an object on the 2D-plane
which is geometrically still a featureless point. We call this principle {\it
hidden feature attachment}. This principle can be extended to other modalities than directions.

\subsection{Qualitative o-point relations\label{basic}}

\noindent In a coarse representation a single o-point induces the
sectors depicted in \nolinebreak{Fig.\ \ref{Regions}}.  
``front,'' 
``back,''
``left,'' and ``right'' 
are linear sectors; 
``left-front,'' ``right-front,'' 
``left-back,'' and ``right-back''
are quadrants.
The position of the point itself is denoted as ``same.''  
This qualitative granularity corresponds to Freksa's single and double
cross calculi
\cite{freksa92b,ScivosN-a:04-finest}.

In ${\cal OPRA}_2$, for the general case where the two points have different positions, we use 
the following relation symbols (the abbreviations ${\rm lf}$,
${\rm lb}$, ${\rm rb}$, ${\rm rf}$ stand for ``left-front,'' 
``left-back,'' ``right-back, '' and ``right-front,'', respectively):\\[-0.3cm]

${}_{\rm front,\,}^{\rm front}$
${}_{\rm front,\,}^{\rm lf}$
${}_{\rm front,\,}^{\rm left}$
${}_{\rm front,\,}^{\rm lb}$
${}_{\rm front,\,}^{\rm back}$
${}_{\rm front,\,}^{\rm rb}$
${}_{\rm front,\,}^{\rm right}$
${}_{\rm front,\,}^{\rm rf}$
${}_{\rm lf\quad ,\,}^{\rm front}$
${}_{\rm lf,\quad\ldots\, ,\quad}^{\rm lf}$
${}_{\rm rf.}^{\rm rf}$\\[-0.3cm]

Here, a qualitative spatial relative direction relation between two o-points
is represented by two pieces of information:
\begin{itemize}
\item the sector (seen from the first o-point) in which the second o-point lies (this
determines the lower part of the relation symbol), and
\item the sector (seen from the second o-point) in which the first o-point lies
(this determines the upper part of the relation symbol).
\end{itemize}

\noindent 
Altogether we obtain $8 \times 8$ base relations for the two points having 
different positions.

\begin{figure}[htb]
\begin{center}
\includegraphics[width=3.5cm]{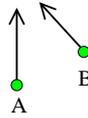}
\caption{\label{Configuration} Qualitative spatial relation between two oriented points at different positions.
The qualitative spatial relation depicted here is $A \; {}_{\rm rf}^{\rm lf} \; B$.}
\end{center}
\end{figure}

\begin{figure}[htb]
\vspace*{-0.4cm}
\begin{center}
\includegraphics[width=2.8cm]{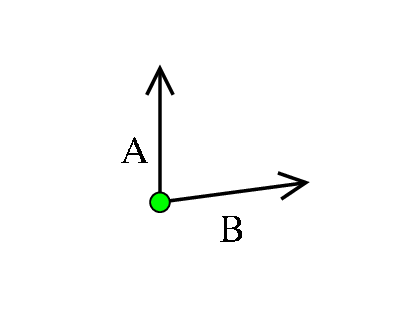}
\vspace*{-0.1cm}\caption{\label{Configuration2} Qualitative spatial relation between two
  oriented points located at the same position.
The qualitative spatial relation depicted here is $A \; {}_{\rm same}^{\rm rf} \; B$.}
\end{center}
\end{figure}

Then the configuration shown in
\nolinebreak{Fig.\ \ref{Configuration}}
is expressed by the relation $A \; {}_{\rm rf}^{\rm lf} \; B$. If both points share the same
position, the lower relation symbol part is the word ``same'' and the upper part denotes
the direction of the second o-point with respect to the first one, as shown in Fig.\ \ref{Configuration2}.

\noindent Altogether we obtain 72 different atomic
relations (eight times eight general relations plus eight with the
o-points at the same position). These relations are jointly exhaustive
and pairwise disjoint (JEPD).  The relation ${}_{\rm same}^{\rm front}$ is the
identity relation.  

\subsection{Elevated o-point relations\label{eopoints}}

\begin{figure}[htb]
\begin{center}
\includegraphics[width=11.5cm]{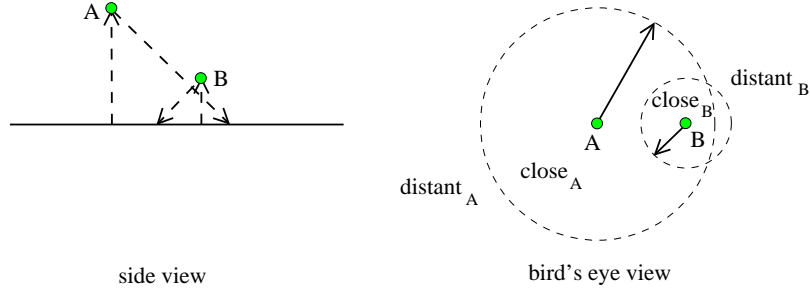}
\caption{\label{elevationFig} Qualitative spatial distance
relation between two oriented points via elevation principle.}
\end{center}
\end{figure}

An extension of binary direction calculi with a local reference direction 
can also be motivated with the following conception.
In this conception local basic perceptions of cognitive agent are the basis
for the agent's ability to distinguish relative distances. In this conception
the point which represents the location is "elevated" above the 2D-plane (see Fig \ref{elevationFig}).
From this observer perspective Gibson's insights about natural perspective
\cite{JamesJGibsonTheEcologicalApproachtoVisualPerception1979} motivate
the availability of  
depth clues which let the observer distinguish local distances based on it's height
as reference distance.
In the case of the \OPRAm calculus in which the entities are oriented points called o-points,
the elevated entities are called elevated o-points, or eo-points.

The granularity depicted in 
Fig. \ref{elevationFig} corresponds to a granularity of $m = 2$
(for the concept of scalable granularity compare \cite{MossakowskiMoratz2010}).
In this granularity we have the following binary distance relations
for points which have different locations:
\begin{displaymath}
{}_{\rm distant \;,}^{\rm distant}\;
{}_{\rm distant \;,}^{\rm equal}\;
{}_{\rm distant \;,}^{\rm close}\;
{}_{\rm equal \;,}^{\rm distant}\;
{}_{\rm equal \;,}^{\rm equal}\;
{}_{\rm equal \;,}^{\rm close}\;
{}_{\rm close \;,}^{\rm distant}\;
{}_{\rm close \;,}^{\rm equal}\;
{}_{\rm close }^{\rm close}\;
\end{displaymath}

The distance relations can be combined with the binary relative direction relations
of the \OPRAm calculus. The relation depicted in Fig. \ref{elevationFig} 
has then the following symbol:
\begin{displaymath}
A \; {}_{\rm rf \; close}^{\rm rf \; distant} \; B
\end{displaymath}

To formally specify the eo-point relations we use two-dimensional
continuous space, in particular ${\mathbb{R}}^2$.
Every eo-point $S$ on the plane
is an ordered triple 
of a point ${\bf p}_S$
represented by its
Cartesian coordinates
$x$ and $y$, with
$x, y \in {\mathbb{R}}$, 
and a direction $\phi_S$, and an internal reference distance $\delta_S$.
The internal reference distance $\delta_S$ is the property which corresponds
to the elevation height in the cognitive motivation.

\begin{displaymath}
S = \left( {\bf p}_S, \phi_S, \delta_S \right) ,   \qquad
{\bf p}_S  = \left( ({\bf p}_S)_x , ({\bf p}_S)_y \right)
\end{displaymath}

The metrical distance between eo-points $A$ and $B$ is their euclidean distance:

\begin{displaymath}
\left| A - B \right| = \sqrt{  
\left( ({\bf p}_A)_x - ({\bf p}_B)_x \right)^2 +
\left( ({\bf p}_A)_y - ({\bf p}_B)_y \right)^2 
}
\end{displaymath}

Then the nine base relations of the binary distance relation in configurations with
${\bf p}_A \not= {\bf p}_B$ are defined in the following way:
\begin{eqnarray*}
\delta_A < \left| A - B \right| > \delta_B   & \iff &    A \; {}_{\rm distant}^{\rm distant} \; B\\
\delta_A < \left| A - B \right| = \delta_B   & \iff &    A \; {}_{\rm distant}^{\rm equal} \; B\\
\delta_A < \left| A - B \right| < \delta_B   & \iff &    A \; {}_{\rm distant}^{\rm close} \; B\\
\delta_A = \left| A - B \right| > \delta_B   & \iff &    A \; {}_{\rm equal}^{\rm distant} \; B\\
\delta_A = \left| A - B \right| = \delta_B   & \iff &    A \; {}_{\rm equal}^{\rm equal} \; B\\
\delta_A = \left| A - B \right| < \delta_B   & \iff &    A \; {}_{\rm equal}^{\rm close} \; B\\
\delta_A > \left| A - B \right| > \delta_B   & \iff &    A \; {}_{\rm close}^{\rm distant} \; B\\
\delta_A > \left| A - B \right| = \delta_B   & \iff &    A \; {}_{\rm close}^{\rm equal} \; B\\
\delta_A > \left| A - B \right| < \delta_B   & \iff &    A \; {}_{\rm close}^{\rm close} \; B
\end{eqnarray*}

\begin{figure}[htb]
\begin{center}
\includegraphics[width=11.5cm]{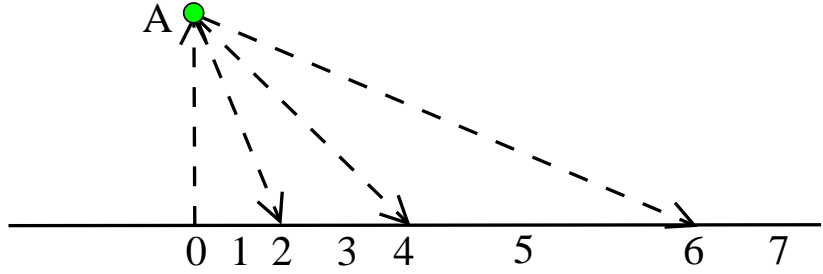}
\caption{\label{EOPRAgranularity} Qualitative spatial distances
with granularty $m = 4$ (e.g. observation angles in $22.5^\circ$ steps).
}
\end{center}
\end{figure}

In OPRAm we have
adjustable granularity with granularity parameter $m$ \cite{MossakowskiMoratz2010}.
Also the distance modality can have an adjustable granularity.
The corresponding schema is demonstrated with granularity $m = 4$ on Fig. \ref{EOPRAgranularity}.

We distinguish the relative distances and
relative directions of the two eo-points $A$ and $B$ expressed by a calculus ${\cal EOPRA}_{m}$
according to the following scheme.
The distance relation symbols $k, l$ are attached to the corresponding direction relation symbols
$i, j$: 
\begin{displaymath}
A \, {\scriptscriptstyle m}\angle_{i|k}^{j|l} \, B
\end{displaymath}

There is also a schema for relative distances established by eo-point pairs.
$A_B$ is an eo-point with $\delta_A = | A - B |$.
The composition of relations follows the standard schema.
As future work the ${\cal EOPRA}_{m}$ composition is will be
computed with condensed semantics, an enumeration of
a small fragment of the domain (this is a broader concept than 
the one used in \cite{oslsa}).
In the case of \EOPRAm , a list of qualitative triangles is generated.
This is a similar apporach like the approach for \OPRAm composition
described in \cite{DBLP:conf/ecai/Moratz06}.

The star calculus can also be augmented with local distances.
Then the same binary distance relations are combined with the absolute direction
relations of the star calculus.

\section{Conclusion}
We presented a first draft of binary position calculi which
combine direction and distance modalities. 
The method which we used is similar to the concept of extending points with an internal
reference direction to get oriented points which are the basic entities
in the \OPRAm calculus. 
Local reference distances which are called elevations are attached to the
basic entities of the calculi.
The principle of attaching references features to a point is called hidden feature
attachment.

\bibliographystyle{plain}
\bibliography{KI05,oslsa}

\end{document}